%% file: paper3008.tex
\RequirePackage{amsmath}
\documentclass[runningheads]{llncs}
\usepackage{xcolor,graphicx,cite,siunitx}

\input{notations.tex}

\begin{document}

\title{Attention-guided Quality Assessment for Automated Cryo-EM Grid Screening
\thanks{Accepted for publication in MICCAI 2020, the 23rd International Conference on Medical Image Computing and Computer Assisted Intervention.}
}

\titlerunning{\smodel}

\author{Hong Xu\inst{1} \and
David E. Timm\inst{2} \and
Shireen Y. Elhabian\inst{1}
}
\authorrunning{Hong Xu, David E. Timm, and Shireen Y. Elhabian}

\institute{Scientific Computing and Imaging Institute, School of Computing, \\University of Utah, Salt Lake City, UT, USA \\
\url{http://www.sci.utah.edu}, \url{https://www.cs.utah.edu} \\
\email{\{hxu,shireen\}@sci.utah.edu} 
\and
Department Of Biochemistry, University of Utah, Salt Lake City, UT, USA
\url{https://medicine.utah.edu/biochemistry} \\
\email{david.timm@biochem.utah.edu}}

\maketitle              

\input{abstract.tex}
\input{introduction.tex}
\input{methods.tex}

\input{results.tex}

\input{conclusion.tex}

\bibliographystyle{splncs04}
\bibliography{references}

\end{document}

%% file: notations.tex
\newcommand{\grid}{\textit{grid}}

\newcommand{\gamt}{12}

\newcommand{\smodel}{\textit{XCryoNet}}

\definecolor{cadmiumorange}{rgb}{0.93, 0.53, 0.18}
\definecolor{celadon}{rgb}{0.67, 0.88, 0.69}
\definecolor{darkolivegreen}{rgb}{0.33, 0.42, 0.18}
\definecolor{darklavender}{rgb}{0.45, 0.31, 0.59}

\renewcommand{\u}{{\bf u}}
\renewcommand{\v}{{\bf v}}

\newcommand{\y}{{\bf y}}

\newcommand{\I}{{\bf I}}

\newcommand{\ben}{\begin{enumerate}}
\newcommand{\een}{\end{enumerate}}

\newcommand{\cmt}[1]{}

%% file: abstract.tex
\begin{abstract}
Cryogenic electron microscopy (cryo-EM) has become an enabling technology in drug discovery and in understanding molecular bases of disease by producing near-atomic resolution (less than 0.4 nm) 3D reconstructions of biological macro-molecules.The imaging process required for 3D reconstructions involves a highly iterative and empirical screening process, starting with the acquisition of low magnification images of the cryo-EM grids. These images are inspected for squares that are likely to contain useful molecular signals. Potentially useful squares within the grid are then imaged at progressively higher magnifications, with the goal of identifying sub-micron areas within circular holes (bounded by the squares) for imaging at high magnification. This arduous, multi-step data acquisition process represents a bottleneck for obtaining a high throughput data collection. Here, we focus on automating the early decision making for the microscope operator, scoring low magnification images of squares, and proposing the first deep learning framework, \smodel, for automated cryo-EM grid screening. \smodel~is a semi-supervised, attention-guided deep learning approach that provides explainable scoring of automatically extracted square images using limited amounts of labeled data. Results show up to 8\% and 37\% improvements over a fully supervised and a no-attention solution, respectively, when labeled data is scarce.
\keywords{Cryo-EM \and Attention models \and Semi-supervised learning}
\end{abstract}

%% file: introduction.tex
\section{Introduction}
\label{sec:intro}

Cryo-electron microscopy (cryo-EM) has recently emerged as an enabling imaging technology for determining 3D structural information of non-crystalline specimens of biologic macromolecules (a.k.a. single particles) at near-atomic resolution (less than 0.4 nm) \cite{new_B_cheng, cryo-AGARD2014113, cryo-state-Ewen, review-Lyumkis2019, Tan15Review}.
Cryo-EM methods are currently the cutting edge of structural biology \cite{new_B_cheng, cryo-AGARD2014113, cryo-state-Ewen, review-Lyumkis2019, Tan15Review}, thanks to recent advances in direct electron detector technology \cite{new_B_cheng} and associated software suites for automating data collection \cite{Tan15Review}, data processing and single particle reconstructions \cite{relion-Scheres2012, new_A_Zheng2017, new_B_cheng, new_C_Punjani2017, new_D_cisTEM, new_E_Tegunov2019}.
Cryo-EM enables highly detailed views of biological machinery (proteins, nucleic acids, and their complexes), which in turn advances the understanding of basic biological systems and mechanisms, furthers the knowledge of the underlying molecular mechanisms of human disease, and provides visual structure-based design of therapeutics for treating human disease \cite{renaud2018cryo,ceska2019cryo}.
Nonetheless, data acquisition alone of a single structure on a state-of-the-art electron microscope costs up to several thousand dollars per day for several days.
Hence, the use of data collection resources should be optimized to yield the highest quality microscopic information for 3D reconstruction.

\begin{figure}
\centering
\includegraphics[width=0.8\textwidth]{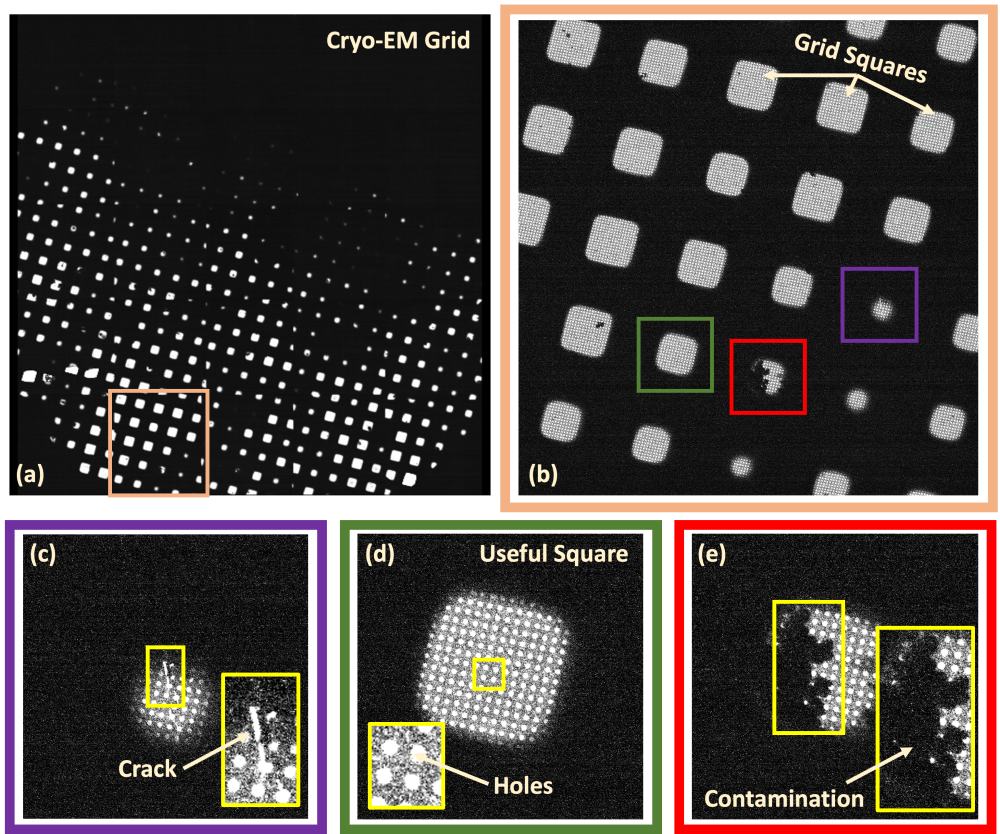}
\caption{
(a) A 3 mm circular grid imaged by the cryo-EM at 135x magnification. Low magnification views of grids are acquired in equally spaced tiles to cover most of the circular grid. (b) One of the 135x tile images that comprises the grid is shown. Bottom: Three exemplar grid squares are shown (i.e., low magnification targets) extracted from the tile illustrating a crack (c), a useful grid square (d) and a square marred by contamination
(e). 
A microscope operator would scan the grid for promising squares to further image at higher magnifications. This process can be slow and imprecise since the operator must manually closely examine squares and squares can be overlooked in the interest of time.
} \label{fig:grid-screening}
\end{figure}

The grid screening process (imaging \& decision making) used to obtain 3D biological information of single particles is a highly iterative, labor intensive process. Screening involves imaging circular grids of fine copper or gold mesh at three or four different levels of magnifications to find the most useful areas for further imaging at the next higher magnification level. 

At the lowest magnifications, cryo-EM grids are manually screened for square-like features of the metal mesh (a.k.a. grid squares or low-magnification targets) that are the most likely to contain useful microscopic signals, using an informal mental \textit{scoring}, to determine which ones should be further imaged at the next magnification. This process is largely based on empirical trial and error \cite{cryo-AGARD2014113}. This \textit{low-magnification target acquisition} process, is illustrated in Fig. \ref{fig:grid-screening}.

Ideal areas for imaging particles within cryo-EM grids are located in thin vitreous ice within the holes (see Fig. \ref{fig:grid-screening}(d)) of the carbon film or gold foil present on the grid surface.
However, targeting holes with enough particle information for the downstream 3D reconstruction task is frequently foiled by the physical damage of fragile grids, excessive amounts of crystalline ice and hydrophobic contaminants, excessively thick ice that is non-transparent or only partially transparent to the electron beam, and/or excessively thin ice that may not accommodate biomolecules or support their native-structures. In particular, at the lowest magnification, a microscope operator manually determines the overall usefulness/quality of a grid square based on visible \textit{attributes}, such as brightness, squareness, cracking, and contamination, that are indicative of such failure at the highest magnification \cite{Tan15Review}. 
However, this arduous, multi-step grid screening process, which entails 3 or 4 multi-scale target acquisition, target scoring, and further imaging subprocesses, represents a bottleneck for obtaining a high throughput of single particle reconstructions \cite{Tan15Review}. Low-magnification target acquisition, in particular, poses significant manual burden since the microscope operator must manually examine squares, increasing the chance of completely overlooking plenty of useful squares in the interest of time. Furthermore, automating low-magnification target acquisition is the backbone process for picking grid squares for higher magnification acquisitions. Such automation paves the way toward a fully automated grid screening process. 
Despite the dramatic impact of the manual burden on imaging throughput, 
automated grid screening in general, and low-magnification target acquisition in particular, are under-explored problems. Most computational work on cryo-EM focuses on the downstream task of reconstructing particles from already collected high magnification images \cite{Tan15Review}. Although existing microscope controller software suites have semi-automated ways of finding cryo-EM squares, these methods depend on operator-defined templates or lattices to identify targets of interest, and use transmittance to determine the viability of said targets \cite{Tan15Review, review-Lyumkis2019, transmittance-holes-LEI200569}. 

A machine learning based solution for automated low-magnification target acquisition is, however, challenging due to the scarcity and associated cost (monetary, manpower, and expertise) of obtaining labeled data and semantic attributes that are manifested at different levels of image scales. Furthermore, explainable automated selection is required for deploying such a solution in practice.

In this paper, we propose the first deep learning based solution, namely \smodel, for explainable, automated grid squares scoring for low-magnification target acquisition. 
To leverage unlabeled data, we borrow ideas from neural network based methods that combine supervised and unsupervised learning by training regularized classifiers using an autoencoder or unsupervised embedding of the data, e.g., \cite{ss-ranzato, ss-kingma, ss-weston}. In particular, we use an autoencoder-like model as a semi-supervised training signal to learn discriminative features from square images that are simultaneously useful for square scoring and reconstruction tasks.
This semi-supervised approach exploits the structure assumption, where grid squares with similar image features are likely to have the same score, by forcing an embedding that captures this structure at the latent space of the autoencoder.
To capture semantic attributes (e.g., cracking and contamination) that are present at different scales, we propose attribute-specific subnetworks that operate on attention-guided input to score a single attribute while learning attention maps that are relevant to that attribute. 

Furthermore, this attention mechanism provides a means of interpreting the resulting scoring via identifying regions in the grid square image that trigger the scoring of a specific attribute.
Attention maps have been used to allow convolutional networks to capture global features relevant to the supervised task beyond the local receptive fields of convolutional filters \cite{Woo_2018_ECCV, Guan2018}. These maps have also been used in the context of interpretable identification of thorax disease \cite{Guan2018}, but under the assumption of a coarse (overall) disease classification that is localized in a single region-of-interest.

Another family of interpretable deep networks obtain attention maps through gradient-based visualization of certain convolutional filters \cite{grad_based_Mah, grad_based_Simon, grad_based_Zeiler}. Nonetheless, such maps are not explicitly learned to reflect attribute-specific interpretations.

We demonstrate that the process of grid screening can be automated in an interpretable way using simple image processing techniques to extract the squares, then using an attention-guided semi-supervised deep network to provide scores representing the quality of said squares.

%% file: methods.tex
\section{Methods}
\label{sec:method}

\begin{figure}
\centering
\includegraphics[width=\textwidth]{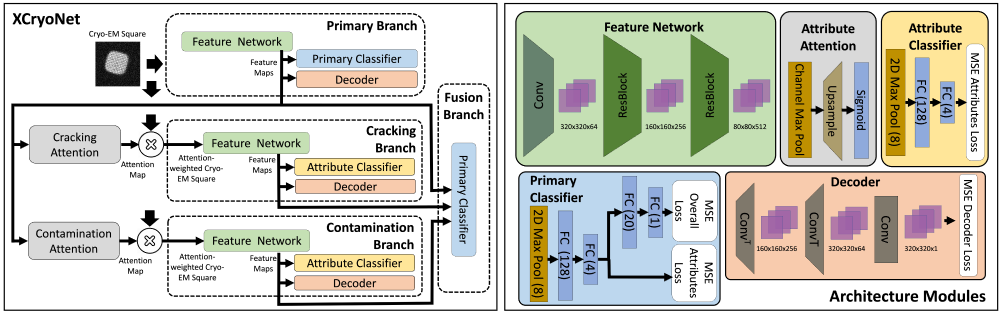}
\caption{\smodel~architecture } \label{fig:architecture}
\end{figure}

The proposed \smodel~architecture, illustrated in Fig. \ref{fig:architecture}, automatically scores low-magnification targets (i.e., squares) on a cryo-EM grid using two levels of granularity. 
Coarse-grained \textit{overall square quality} reflects the perceived overall quality of vitreous ice in a grid square.
Fine-grained visible \textit{attributes} (e.g., brightness, squareness, cracking, and contamination) are specific abstract image qualities visible at low magnification indicative of loss of potentially informative microscopic signal at higher magnification levels for 3D reconstructions. 

\smodel~consists of three types of interacting subnetworks (or branches) that are trained end-to-end. First, the \textit{primary} branch aims at solving the primary scoring task for both coarse- and fine-grained qualities. Second, the \textit{attribute} branch aims at solving the scoring task of an individual fine-grained attribute. Third, the \textit{fusion} branch combines features learned from the primary and the attributes branches (via the feature networks) to solve the primary scoring task. Such fusion aggregates features learned by the primary and attribute branches to boost the performance of the primary scoring task. Hereafter, we present the motivations and design choices of these interacting branches.

\vspace{0.05in}
\noindent\textbf{Attributes and Labeling.} Brightness concerns the overall intensity of the square. Squareness is defined by how much the image resembles a square. Cracking is determined by the portion of the surface that has fissures. Contamination is a measure of the portion of the surface covered by artifacts. We encode these attributes into a vector
$\y = [y_{b},y_{s},y_{cr},y_{co},y_{o}]$, where $y_{b},y_{s},y_{cr},y_{co}$, and $y_{o}$ denote the score of the brightness, squareness, cracking, contamination, and overall quality, respectively and the score $y_* \in \{0,1,2,3,4\}$.

\vspace{0.05in}
\noindent\textbf{Primary Branch.} The objective of the primary branch is to learn from the global image characteristics to make an informed decision. This branch consists of a feature network, a primary classifier network and a decoder network. 
The feature network has a convolutional layer and two ResBlocks \cite{resnet}. The primary classifier network learns an explicit nonlinear functional mapping that infers the overall score directly from the attributes to enforce the dependency of the overall score on the fine-grained attributes. 
It consists of a pooling layer, two fully connected layers for attribute regression, and two fully connected layers for overall score regression. 

The decoder network is added after the second ResBlock to account for the scarcity of labeled data by enforcing discriminative features for the scoring task while also being useful for the input reconstruction task. It is comprised of two transpose-convolution layers and one convolution layer.

The primary network is trained by minimizing a supervised loss, $\mathcal{L}_{S}^p$, that combines the attributes loss, the overall quality loss, and an unsupervised loss, $\mathcal{L}_{U}^p$, for input reconstruction via the decoder. 
\begin{equation}
\mathcal{L}^p(\Theta_p) =  \mathcal{L}_{S}^p(\Theta_p) + \mathcal{L}_{U}^p(\Theta_p)
\end{equation}
\noindent The supervised attribute loss is defined by
\begin{equation}
\mathcal{L}_{S}^p(\Theta_p) = \text{MSE}([y_{b},y_{s},y_{cr},y_{co}], [\hat{y}_b,\hat{y}_s,\hat{y}_{cr},\hat{y}_{co}]) + \text{MSE}([y_o], [\hat{y_o}]).
\end{equation}
\noindent where $\Theta_p$ are the parameters of the primary network, $\hat{y}_*$ is the prediction for the score value of the $y_*$, and $\text{MSE}(\u,\v)$ is the mean square error between elements of $\u$ and $\v$. The decoder loss is defined by
\begin{equation}
\mathcal{L}_{U}^p(\Theta_p) =  \text{MSE}(\I, \hat{\I})
\end{equation}
\noindent where $\I$ is the input grid square image and $\hat{\I}$ is the reconstructed image.

\vspace{0.05in}
\noindent\textbf{Attention Guidance.} 
The primary branch is able to infer global scale attributes (e.g., brightness and squareness), but fails to score attributes with multi-scale presence (e.g., cracking and contamination) in a meaningful manner. Feeding attention-guided squares to attribute branches mitigates the poor cracking and contamination scores by dedicating two subnetworks, the cracking branch and the contamination branch, to the task of scoring individual fine-grained attributes from attention-weighted inputs.
The attention-weighted squares are generated by taking the output feature maps from the feature network of the primary branch and distributing the channels evenly among every attribute. In particular, we feed half of the channels to the cracking attention and half to the contamination attention. This separation allows the primary feature network to learn attribute-specific features that are relevant to generating attention maps for each attribute.
Not only does this separation produce different feature maps for each branch, but it also allows the attribute branches to serve as regularizers for the primary network to learn to focus on finding the relevant attribute-specific features. 
Attribute-specific attention-weighted squares are then generated by channel-wise max-pooling the channels corresponding to each attribute, up-sampling to the input size to match the grid square dimension for attention guidance, and a sigmoid function to force a $(0,1)-$range. The attention-guided grid squares to be fed to the attribute branches are obtained by multiplying the attention map by the grid square image to highlight relevant regions for scoring that attribute.

\vspace{0.05in}
\noindent\textbf{Attribute Branch.} The objective of the attribute branch is to focus on scoring an individual attribute by focusing on areas highlighted by the attention guidance. Attribute branches share a similar architectural design to the primary branch, but instead of regressing on all the attributes, they regresses on a single one. The input of these branches are the attention-weighted grid squares obtained from the primary branch and the attribute attention, and each attribute branch is expected to reconstruct its attention-weighted input using its decoder for semi-supervised learning. 

Consider the attribute branch for inferring $y_*$ and let $\I_*$ be its attention-weighted input. Similar to the primary branch loss, the attribute branch is trained using a combination of supervised and unsupervised losses.
\begin{equation}
\mathcal{L}^*(\Theta_*) =  \mathcal{L}_{S}^*(\Theta_*) + \mathcal{L}_{U}^*(\Theta_*)
\end{equation}
\noindent where $\mathcal{L}_{S}^*(\Theta_*)$ is the mean square error between $y_*$ and $\hat{y}_*$, and $\mathcal{L}_{U}^*(\Theta_*)$ is the mean square error between $\I_*$ and $\hat{\I}_*$

\vspace{0.05in}
\noindent\textbf{Fusion Branch.} The fusion branch combines the feature maps obtained from the primary branch as well as the attribute branches to make a final prediction that leverage both global and multi-scale features.
The fusion branch's loss $\mathcal{L}^f(\Theta_f)$ is identical to the supervised loss of the primary branch, $\mathcal{L}^p$.

\vspace{0.05in}
\noindent\textbf{\smodel~Training.}
The training procedure is dissected into three alternating steps.training.

(1) \textit{Primary and attribute training.} The feature, primary/attribute classifiers, and the primary decoder networks are trained by minimizing the supervised losses ($\mathcal{L}_{S}^p(\Theta_p)$ and $\mathcal{L}_{S}^*(\Theta_*)$), and the primary decoder loss $\mathcal{L}_{U}^p(\Theta_p)$. 
(2) \textit{Attribute autoencoder training.} This procedure freezes the parameters of the whole network except for the encoder (feature) network and decoder network of the attribute branches, and uses the attribute decoder loss $\mathcal{L}_{U}^*(\Theta_*)$ to back-propagate.

The purpose of separating (1) from (2) is such that the decoder output does not influence the construction of the attention-weighted squares.

(3) \textit{Fusion training.} Finally, the fusion network parameters are isolated and trained using the fusion loss $\mathcal{L}^f(\Theta_f)$. We train this separately as to properly isolate the individual attribute branches from learning from other attributes.

%% file: results.tex
\begin{figure}
\centering
\includegraphics[width=0.95\textwidth]{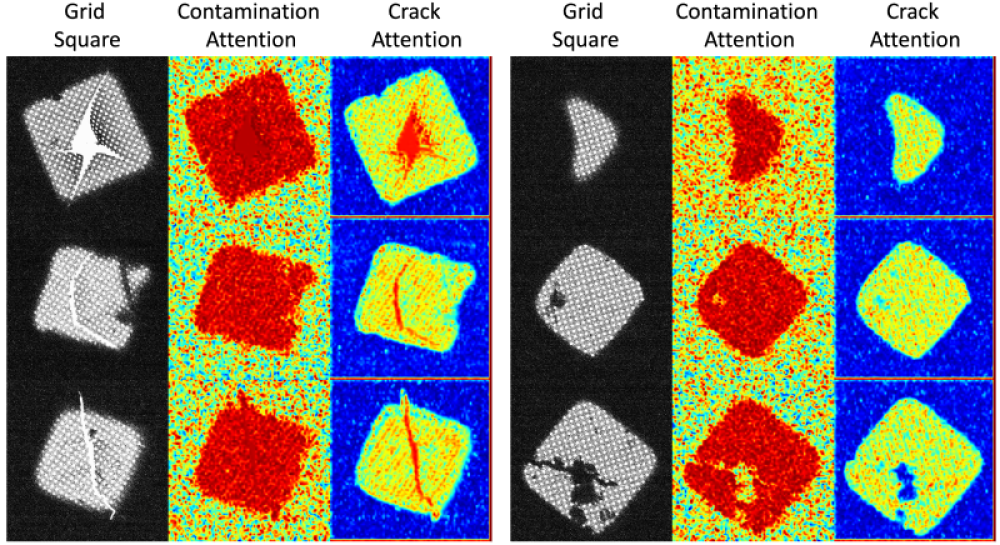}
\caption{Attention maps at epoch 75 for \smodel~with 900 labeled samples and 1500 unlabeled.
} \label{fig:heat-maps}
\end{figure}

\section{Results}
\label{sec:experiments}

Our experiments focus on comparing the semi-supervised versus the fully supervised setting for the primary branch (i.e., no attention guidance) and the full \smodel.
A fitting performance metric that allows a quantitative measure of the proximity of the predicted score to the true score is the mean absolute difference between the true scores and the predicted scores $d(y_*, \hat{y_*}) = \frac{1}{n}\sum_{i=0}^{n}|y_* - \hat{y_*}|$. We report the quantitative performance metrics of the various settings and show qualitative results in the form of the attention maps.

\vspace{0.05in}
\noindent\textbf{Dataset and Preprocessing.} \label{subsec:experiments:preprocessing}
The input we work with are \gamt\ MRC/CCP4 2014 files (standard files for cryo-EM image/movie files) along with the microscope parameter files that Thermo-Fisher's EPU software outputs that are used for stitching the individual tiles to fit into a $5 \times 5$ montage.
The extraction of squares relies on a normalized cross-correlation based template matching with a custom template created according to the pixel intensity distribution of the grid.  
We acquire 250 $640 \times 640$ images of squares per \grid, totaling about 3000 for the \gamt\ \grid s. 
The brightness scores of the extracted squares are set to the mean pixel value of non-zero valued pixels scaled to a score value. The squareness is obtained by applying canny edge detection, then dividing the non-zero pixel area over the total area of a minimum area square scaled to a score value. Finally, the cracking and contamination scores are manually labeled by an experienced microscope operator.
These are the squares that are fed to the \smodel.

\begin{table}
\caption{The quantitative measure of score proximity (lower is better) on held-out (testing) grid squares of the fully supervised (FS) and semi-supervised (SS) versions of the primary and \smodel~with different amount of labeled examples used to train the model. Each network is trained four times for 75 epochs, which were enough for convergence, with random uniformly selected training and test samples; the means and standard deviations among runs are reported.
}\label{table:results}
\begin{tabular}{|l|p{17mm}| |l|l|l|p{17mm}| |l|}
\hline
Method & Supervision$|$ \newline \#Labeled$|$ \newline \#Unlabeled & Brightness & Squareness & Cracking & Contami- \newline nation & Overall\\
\hline
Primary & SS$|100|1500$ & $1.54\pm.418$          & $1.85\pm.309$          & $1.88\pm.363$          & $1.93\pm.148$          & $2.18\pm.174$ \\
Primary & FS$|100|0$ &    $2.15\pm.719$          & $1.49\pm.265$          & $1.81\pm.061$          & $1.63\pm.394$          & $2.57\pm.443$ \\
\smodel & SS$|100|1500$ & $\mathbf{0.91\pm.295}$ & $\mathbf{1.08\pm.245}$ & $\mathbf{1.35\pm.143}$ & $\mathbf{1.23\pm.151}$ & $\mathbf{1.38\pm.386}$ \\
\smodel & FS$|100|0$ &    $1.01\pm.290$          & $1.23\pm.442$          & $1.46\pm.154$          & $1.31\pm.567$          & $1.50\pm.340$\\
\hline
Primary & SS$|500|1500$ & $\mathbf{0.26\pm.010}$ & $\mathbf{0.53\pm.022}$ & $0.95\pm.061$          & $0.64\pm.022$          & $0.62\pm.032$\\
Primary & FS$|500|0$    & $0.30\pm.025$          & $0.54\pm.021$          & $\mathbf{0.86\pm.055}$ & $\mathbf{0.62\pm.037}$ & $0.62\pm.022$ \\
\smodel & SS$|500|1500$ & $0.28\pm.028$          & $0.53\pm.059$          & $0.91\pm.057$          & $0.66\pm.029$          & $0.58\pm.059$\\
\smodel & FS$|500|0$    & $0.32\pm.034$          & $0.57\pm.033$          & $1.00\pm.064$          & $0.75\pm.043$          & $\mathbf{0.58\pm.025}$\\
\hline
Primary & SS$|900|1500$ & $\mathbf{0.26\pm.024}$ & $0.51\pm.035$          & $0.86\pm.053$          & $0.64\pm.036$          & $0.52\pm.036$\\
Primary & FS$|900|0$    & $0.31\pm.021$          & $\mathbf{0.49\pm.027}$ & $\mathbf{0.71\pm.033}$ & $\mathbf{0.62\pm.011}$ & $\mathbf{0.47\pm.019}$\\
\smodel & SS$|900|1500$ & $0.36\pm.051$          & $0.55\pm.033$          & $0.87\pm.061$          & $0.62\pm.057$          & $0.51\pm.005$\\
\smodel & FS$|900|0$    & $0.45\pm.097$          & $0.66\pm.133$          & $1.40\pm.399$          & $0.79\pm.146$          & $0.83\pm.234$\\
\hline
\end{tabular}
\end{table}

\vspace{0.05in}
\noindent\textbf{Quantitative and Qualitative Results.}
\label{subsec:experiments:results}

Table \ref{table:results} reports $d(y_*, \hat{y_*})$ for coarse- and fine-grained attributes for fully and semi-supervised settings with and without attention guidance. The experiments were run on an Intel® Core™ i7-6850K @ 3.60GHz x 12 64GB DDR4 machine with a GTX 1080 Ti GPU. The fully-supervised primary branch takes 11 minutes to train for 75 epochs on 100 labeled examples, whereas the semi-supervised primary branch takes an hour and a half with 1500 additional unlabeled examples. In general, the semi-supervised (with 1500 unlabeled examples) runs take from 3 (with 900 labeled examples) to 8 (with 100 labeled examples) times longer than their fully-supervised counterparts. \smodel~takes 4 to 5 times longer to train than just running the primary branch. Test time is almost instantaneous, thanks to the feed-forward architecture of \smodel. Results shows that as the ratio between labeled and unlabeled data increases, or the amount of labeled signal sufficiently informs the classifier and feature networks, the effect of semi-supervision diminishes. Likewise, the semi-supervised \smodel~can significantly outperform the primary-only setting the scarcer the labeled data becomes. 

Fig. \ref{fig:heat-maps} shows examples of generated attention maps for the cracking and contamination attribute branches. These attention maps are able to identify most instances of heavy cracking and contamination, but still struggle to detect more subtle ones. The contamination attention maps highlights the portions of the square without contamination, while the cracking ones highlight the cracks themselves. This is because any dark area within a grid square (which are all supposed to be the same size) is considered to be contamination, so the network must focus on the portion of the grid that is not contaminated to score the contamination attribute accurately.

%% file: conclusion.tex
\section{Conclusion}
\label{sec:conclusion}

We have presented \smodel, a semi-supervised, attention-guided deep learning approach that provides interpretable scoring of automatically extracted cryo-EM grid squares using limited amounts of labeled data.
Results show that trained \smodel s are able to mimic the mental scoring process of a microscope operator, providing both interpretable attention maps and good scoring performance, even with scarce labeled data.
This work represents the first step in fully automating the grid screening process for cryo-EM, which will significantly increase the throughput of high quality reconstructions without the need to waste valuable man-power and research funds.